# Semi-supervised User Geolocation via Graph Convolutional Networks

**Afshin Rahimi**     **Trevor Cohn**     **Timothy Baldwin**
School of Computing and Information Systems
The University of Melbourne
arahimi@student.unimelb.edu.au
{t.cohn,tbaldwin}@unimelb.edu.au

## Abstract

Social media user geolocation is vital to many applications such as event detection. In this paper, we propose GCN, a multiview geolocation model based on Graph Convolutional Networks, that uses both text and network context. We compare GCN to the state-of-the-art, and to two baselines we propose, and show that our model achieves or is competitive with the state-of-the-art over three benchmark geolocation datasets when sufficient supervision is available. We also evaluate GCN under a minimal supervision scenario, and show it outperforms baselines. We find that highway network gates are essential for controlling the amount of useful neighbourhood expansion in GCN.

## 1 Introduction

User geolocation, the task of identifying the "home" location of a user, is an integral component of many applications ranging from public health monitoring (Paul and Dredze, 2011; Chon et al., 2015; Yepes et al., 2015) and regional studies of sentiment, to real-time emergency awareness systems (De Longueville et al., 2009; Sakaki et al., 2010), which use social media as an implicit information resource about people.

Social media services such as Twitter rely on IP addresses, WiFi footprints, and GPS data to geolocate users. Third-party service providers don't have easy access to such information, and have to rely on public sources of geolocation information such as the profile location field, which is noisy and difficult to map to a location (Hecht et al., 2011), or geotagged tweets, which are publicly available for only 1% of tweets (Cheng et al., 2010; Morstatter et al., 2013). The scarcity of publicly available location information motivates predictive user geolocation from information such as tweet text and social interaction data.

Most previous work on user geolocation takes the form of either supervised text-based approaches (Wing and Baldridge, 2011; Han et al., 2012) relying on the geographical variation of language use, or graph-based semi-supervised label propagation relying on location homophily in user–user interactions (Davis Jr et al., 2011; Jurgens, 2013).

Both text and network views are critical in geolocating users. Some users post a lot of local content, but their social network is lacking or is not representative of their location; for them, text is the dominant view for geolocation. Other users have many local social interactions, and mostly use social media to read other people's comments, and for interacting with friends. Single-view learning would fail to accurately geolocate these users if the more information-rich view is not present. There has been some work that uses both the text and network views, but it either completely ignores unlabelled data (Li et al., 2012a; Miura et al., 2017), or just uses unlabelled data in the network view (Rahimi et al., 2015b; Do et al., 2017). Given that the 1% of geotagged tweets is often used for supervision, it is crucial for geolocation models to be able to leverage unlabelled data, and to perform well under a minimal supervision scenario.

In this paper, we propose GCN, an end-to-end user geolocation model based on Graph Convolutional Networks (Kipf and Welling, 2017) that jointly learns from text and network information to classify a user timeline into a location. Our contributions are: (1) we evaluate our model under a minimal supervision scenario which is close to real world applications and show that GCN outperforms two strong baselines; (2) given sufficient supervision, we show that GCN is competitive, although the much simpler MLP-TXT+NET outper-forms state-of-the-art models; and (3) we show that highway gates play a significant role in controlling the amount of useful neighbourhood smoothing in GCN.[1]

## 2 Model

We propose a transductive multiview geolocation model, GCN, using Graph Convolutional Networks ("GCN": Kipf and Welling (2017)). We also introduce two multiview baselines: MLP-TXT+NET based on concatenation of text and network, and DCCA based on Deep Canonical Correlation Analysis (Andrew et al., 2013).

### 2.1 Multiview Geolocation

Let $X \in \mathbb{R}^{|U| \times |V|}$ be the text view, consisting of the bag of words for each user in $U$ using vocabulary $V$, and $A \in \mathbb{1}^{|U| \times |U|}$ be the network view, encoding user–user interactions. We partition $U = U_S \cup U_H$ into a supervised and heldout (unlabelled) set, $U_S$ and $U_H$, respectively. The goal is to infer the location of unlabelled samples $Y_U$, given the location of labelled samples $Y_S$, where each location is encoded as a one-hot classification label, $y_i \in \mathbb{1}^c$ with $c$ being the number of target regions.

### 2.2 GCN

GCN defines a neural network model $f(X, A)$ with each layer:

$$\hat{A} = \tilde{D}^{-\frac{1}{2}}(A + \lambda I)\tilde{D}^{-\frac{1}{2}}$$
$$H^{(l+1)} = \sigma\left(\hat{A}H^{(l)}W^{(l)} + b\right), \quad (1)$$

where $\tilde{D}$ is the degree matrix of $A + \lambda I$; hyperparameter $\lambda$ controls the weight of a node against its neighbourhood, which is set to 1 in the original model (Kipf and Welling, 2017); $H^0 = X$ and the $d_{\text{in}} \times d_{\text{out}}$ matrix $W^{(l)}$ and $d_{\text{out}} \times 1$ matrix $b$ are trainable layer parameters; and $\sigma$ is an arbitrary nonlinearity. The first layer takes an average of each sample and its immediate neighbours (labelled and unlabelled) using weights in $\hat{A}$, and performs a linear transformation using $W$ and $b$ followed by a nonlinear activation function ($\sigma$). In other words, for user $u_i$, the output of layer $l$ is computed by:

$$\vec{h}_i^{l+1} = \sigma\left(\sum_{j \in \text{nhood}(i)} \hat{A}_{ij}\vec{h}_j^l W^l + b^l\right), \quad (2)$$

[1] Code and data available at https://github.com/afshinrahimi/geographconv

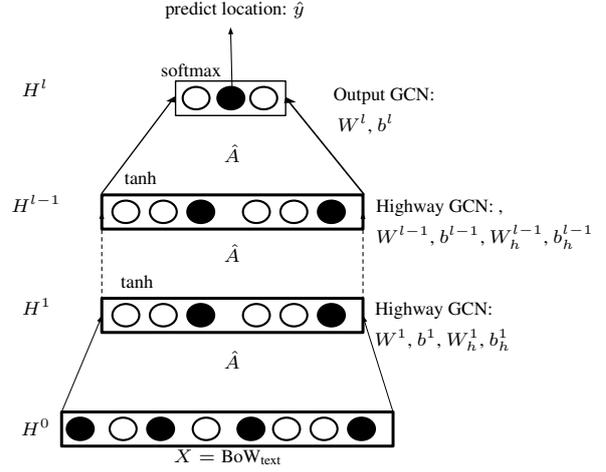

Figure 1: The architecture of GCN geolocation model with layer-wise highway gates ($W_h^i$, $b_h^i$). GCN is applied to a BoW model of user content over the @-mention graph to predict user location.

where $W^l$ and $b^l$ are learnable layer parameters, and $\text{nhood}(i)$ indicates the neighbours of user $u_i$. Each extra layer in GCN extends the neighbourhood over which a sample is smoothed. For example a GCN with 3 layers smooths each sample with its neighbours up to 3 hops away, which is beneficial if location homophily extends to a neighbourhood of this size.

#### 2.2.1 Highway GCN

Expanding the neighbourhood for label propagation by adding multiple GCN layers can improve geolocation by accessing information from friends that are multiple hops away, but it might also lead to propagation of noisy information to users from an exponentially increasing number of expanded neighbourhood members. To control the required balance of how much neighbourhood information should be passed to a node, we use layer-wise gates similar to highway networks. In highway networks (Srivastava et al., 2015), the output of a layer is summed with its input with gating weights $T(\vec{h}^l)$:

$$T(\vec{h}^l) = \sigma\left(W_t^l \vec{h}^l + b_t^l\right)$$
$$\vec{h}^{l+1} = \vec{h}^{l+1} \circ T(\vec{h}^l) + \vec{h}^l \circ (1 - T(\vec{h}^l)), \quad (3)$$

where $\vec{h}^l$ is the incoming input to layer $l + 1$, ($W_t^l, b_t^l$) are gating weights and bias variables, $\circ$ is elementwise multiplication, and $\sigma$ is the Sigmoid function.

## 2.3 DCCA

Given two views $X$ and $\hat{A}$ (from Equation 1) of data samples, CCA (Hotelling, 1936), and its deep version (DCCA) (Andrew et al., 2013) learn functions $f_1(X)$ and $f_2(\hat{A})$ such that the correlation between the output of the two functions is maximised:

$$\rho = \text{corr}(f_1(X), f_2(\hat{A})). \quad (4)$$

The resulting representations of $f_1(X)$ and $f_2(\hat{A})$ are the compressed representations of the two views where the uncorrelated noise between them is reduced. The new representations ideally represent user communities for the network view, and the language model of that community for the text view, and their concatenation is a multiview representation of data, which can be used as input for other tasks.

In DCCA, the two views are first projected to a lower dimensionality using a separate multilayer perceptron for each view (the $f_1$ and $f_2$ functions of Equation 4), the output of which is used to estimate the CCA cost:

$$\begin{aligned} \text{maximise:} \quad & \text{tr}(W_1^T \Sigma_{12} W_2) \\ \text{subject to:} \quad & W_1^T \Sigma_{11} W_1 = W_2^T \Sigma_{22} W_2 = I \end{aligned} \quad (5)$$

where $\Sigma_{11}$ and $\Sigma_{22}$ are the covariances of the two outputs, and $\Sigma_{12}$ is the cross-covariance. The weights $W_1$ and $W_2$ are the linear projections of the MLP outputs, which are used in estimating the CCA cost. The optimisation problem is solved by SVD, and the error is backpropagated to train the parameters of the two MLPs and the final linear projections. After training, the two networks are used to predict new projections for unseen data. The two projections of unseen data — the outputs of the two networks — are then concatenated to form a multiview sample representation, as shown in Figure 2.

## 3 Experiments

### 3.1 Data

We use three existing Twitter user geolocation datasets: (1) GEOTEXT (Eisenstein et al., 2010), (2) TWITTER-US (Roller et al., 2012), and (3) TWITTER-WORLD (Han et al., 2012). These datasets have been used widely for training and evaluation of geolocation models. They are all pre-partitioned into training, development and test sets. Each user is represented by the concatenation of their tweets, and labelled with the latitude/longitude of the first collected geotagged tweet in the case of GEOTEXT and TWITTER-US, and the centre of the closest city in the case of TWITTER-WORLD. GEOTEXT and TWITTER-US cover the continental US, and TWITTER-WORLD covers the whole world, with 9k, 449k and 1.3m users, respectively. The labels are the discretised geographical coordinates of the training points using a $k$-d tree following Roller et al. (2012), with the number of labels equal to 129, 256, and 930 for GEOTEXT, TWITTER-US, and TWITTER-WORLD, respectively.

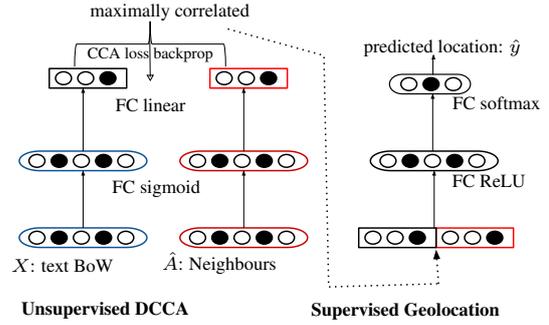

Figure 2: The DCCA model architecture: First the two text and network views $X$ and $\hat{A}$ are fed into two neural networks (left), which are unsupervisedly trained to maximise the correlation of their outputs; next the outputs of the networks are concatenated, and fed as input to another neural network (right), which is trained supervisedly to predict locations.

### 3.2 Constructing the Views

We build matrix $\hat{A}$ as in Equation 1 using the collapsed @-mention graph between users, where two users are connected ($A_{ij} = 1$) if one mentions the other, or they co-mention another user. The text view is a BoW model of user content with binary term frequency, inverse document frequency, and $l_2$ normalisation of samples.

### 3.3 Model Selection

For GCN, we use highway layers to control the amount of neighbourhood information passed to a node. We use 3 layers in GCN with size 300, 600, 900 for GEOTEXT, TWITTER-US and TWITTER-WORLD respectively. Note that the final softmax layer is also graph convolutional, which sets the radius of the averaging neighbourhood to 4. The

$k$-d tree bucket size hyperparameter which controls the maximum number of users in each cluster is set to 50, 2400, and 2400 for the respective datasets, based on tuning over the validation set. The architecture of `GCN-LP` is similar, with the difference that the text view is set to zero. In `DCCA`, for the unsupervised networks we use a single sigmoid hidden layer with size 1000 and a linear output layer with size 500 for the three datasets. The loss function is CCA loss, which maximises the output correlations. The supervised multilayer perceptron has one hidden layer with size 300, 600, 1000 for GEOTEXT, TWITTER-US, and TWITTER-WORLD, respectively, which we set by tuning over the development sets. We evaluate the models using Median error, Mean error, and Acc@161, accuracy of predicting a user within 161km or 100 miles from the known location.

### 3.4 Baselines

We also compare `DCCA` and `GCN` with two baselines:

`GCN-LP` is based on `GCN`, but for input, instead of text-based features, we use one-hot encoding of a user's neighbours, which are then convolved with their $k$-hop neighbours using the `GCN`. This approach is similar to label propagation in smoothing the label distribution of a user with that of its neighbours, but uses graph convolutional networks which have extra layer parameters, and also a gating mechanism to control the smoothing neighbourhood radius. Note that for unlabelled samples, the predicted labels are used for input after training accuracy reaches 0.2.

`MLP-TXT+NET` is a simple transductive supervised model based on a single layer multilayer perceptron where the input to the network is the concatenation of the text view $X$, the user content's bag-of-words and $\hat{A}$ (Equation 1), which represents the network view as a vector input. For the hidden layer we use a ReLU nonlinearity, and sizes 300, 600, and 600 for GEOTEXT, TWITTER-US, and TWITTER-WORLD, respectively.

## 4 Results and Analysis

### 4.1 Representation

Deep CCA and GCN are able to provide an unsupervised data representation in different ways. Deep CCA takes the two text-based and network-based views, and finds deep non-linear transformations that result in maximum correlation between the two views (Andrew et al., 2013). The representations can be visualised using t-SNE, where we hope that samples with the same label are clustered together. GCN, on the other hand, uses graph convolution. The representations of 50 samples from each of 4 randomly chosen labels of GEOTEXT are shown in Figure 3. As shown, Deep CCA seems to slightly improve the representations from pure concatenation of the two views. GCN, on the other hand, substantially improves the representations. Further application of GCN results in more samples clumping together, which might be desirable when there is strong homophily.

### 4.2 Labelled Data Size

To achieve good performance in supervised tasks, often large amounts of labelled data are required, which is a big challenge for Twitter geolocation, where only a small fraction of the data is geotagged (about 1%). The scarcity of supervision indicates the importance of semi-supervised learning where unlabelled (e.g. non-geotagged) tweets are used for training. The three models we propose (`MLP-TXT+NET`, `DCCA`, and `GCN`) are all transductive semi-supervised models that use unlabelled data, however, they are different in terms of how much labelled data they require to achieve acceptable performance. Given that in a real-world scenario, only a small fraction of data is geotagged, we conduct an experiment to analyse the effect of labelled samples on the performance of the three geolocation models. We provided the three models with different fractions of samples that are labelled (in terms of % of dataset samples) while using the remainder as unlabelled data, and analysed their Median error performance over the development set of GEOTEXT, TWITTER-US, and TWITTER-WORLD. Note that the text and network view, and the development set, remain fixed for all the experiments. As shown in Figure 4, when the fraction of labelled samples is less than 10% of all the samples, `GCN` and `DCCA` outperform `MLP-TXT+NET`, as a result of having fewer parameters, and therefore, lower supervision requirement to optimise them. When enough training data is available (e.g. more than 20% of all the samples), `GCN` and `MLP-TXT+NET` clearly outperform `DCCA`, possibly as a result of directly modelling the interactions between network and text views. When all the training samples of the two larger datasets (95% and 98% for TWITTER-US and TWITTER-WORLD, respectively) are available to the models, `MLP-TXT+NET` outperforms `GCN`. Note that the number of parameters increases from `DCCA` to `GCN` and to `MLP-TXT+NET`. In 1% for GEOTEXT, `DCCA` outperforms `GCN` as a result of having fewer parameters and just a few labelled samples, insufficient to train the parameters of `GCN`.

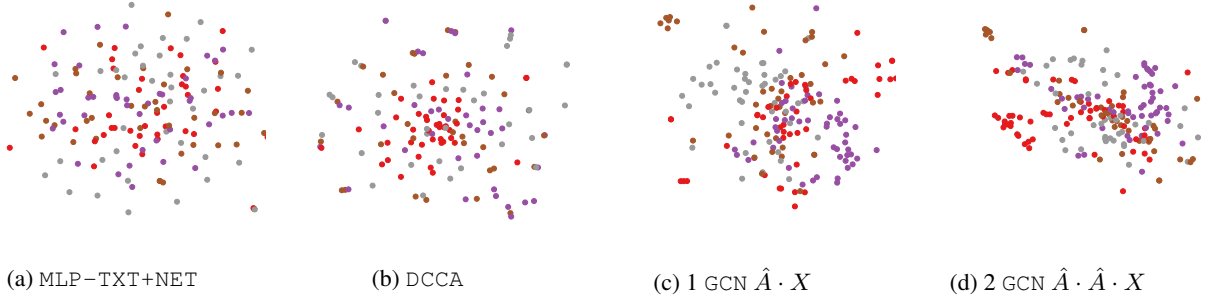

(a) `MLP-TXT+NET`  (b) `DCCA`  (c) 1 `GCN` $\hat{A} \cdot X$  (d) 2 `GCN` $\hat{A} \cdot \hat{A} \cdot X$

Figure 3: Comparing t-SNE visualisations of 50 training samples from each of 4 randomly chosen regions of GEOTEXT using various data representations: (a) concatenation of $\hat{A}$ (Equation 1); (b) concatenation of `DCCA` transformation of text-based and network-based views $X$ and $\hat{A}$; (c) applying graph convolution $\hat{A} \cdot X$; and (d) applying graph convolution twice $\hat{A} \cdot \hat{A} \cdot X$

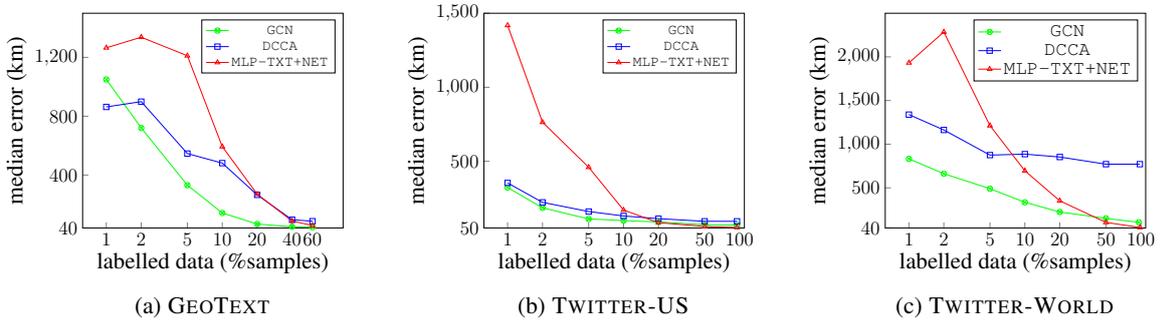

(a) GEOTEXT  (b) TWITTER-US  (c) TWITTER-WORLD

Figure 4: The effect of the amount of labelled data available as a fraction of all samples for GEOTEXT, TWITTER-US, and TWITTER-WORLD on the development performance of `GCN`, `DCCA`, and `MLP-TXT+NET` models in terms of Median error. The dataset sizes are 9k, 440k, and 1.4m for the three datasets, respectively.

### 4.3 Highway Gates

Adding more layers to `GCN` expands the graph neighbourhood within which the user features are averaged, and so might introduce noise, and consequently decrease accuracy as shown in Figure 5 when no gates are used. We see that by adding highway network gates, the performance of `GCN` slightly improves until three layers are added, but then by adding more layers the performance doesn't change that much as gates are allowing the layer inputs to pass through the network without much change. The performance peaks at 4 layers which is compatible with the distribution of shortest path lengths shown in Figure 6.

### 4.4 Performance

The performance of the three proposed models (`MLP-TXT+NET`, `DCCA` and `GCN`) is shown in Table 1. The models are also compared with supervised text-based methods (Wing and Baldridge, 2014; Cha et al., 2015; Rahimi et al., 2017b), a network-based method (Rahimi et al., 2015a) and `GCN-LP`, and also joint text and network models (Rahimi et al., 2017b; Do et al., 2017; Miura et al., 2017). `MLP-TXT+NET` and `GCN` outperform all the text- or network-only models, and also the hybrid model of Rahimi et al. (2017b), indicating that joint modelling of text and network features is important. `MLP-TXT+NET` is competitive with Do et al. (2017), outperforming it on larger datasets, and underperforming on GEO-

|  | GEOTEXT | | | TWITTER-US | | | TWITTER-WORLD | | |
|---|---|---|---|---|---|---|---|---|---|
|  | Acc@161↑ | Mean↓ | Median↓ | Acc@161↑ | Mean↓ | Median↓ | Acc@161↑ | Mean↓ | Median↓ |
| Text (inductive) | | | | | | | | | |
| Rahimi et al. (2017b) | 38 | 844 | 389 | 54 | 554 | 120 | 34 | 1456 | 415 |
| Wing and Baldridge (2014) | — | — | — | 48 | 686 | 191 | 31 | 1669 | 509 |
| Cha et al. (2015) | — | 581 | 425 | — | — | — | — | — | — |
| Network (transductive) | | | | | | | | | |
| Rahimi et al. (2015a) | 58 | 586 | 60 | 54 | 705 | 116 | 45 | 2525 | 279 |
| GCN-LP | 58 | 576 | 56 | 53 | 653 | 126 | 45 | 2357 | 279 |
| Text+Network (transductive) | | | | | | | | | |
| Do et al. (2017) | **62** | **532** | **32** | **66** | 433 | **45** | 53 | 1044 | 118 |
| Miura et al. (2017) | — | — | — | 61 | 481 | 65 | — | — | — |
| Rahimi et al. (2017b) | 59 | 578 | 61 | 61 | 515 | 77 | 53 | 1280 | 104 |
| MLP-TXT+NET | 58 | 554 | 58 | **66** | 420 | 56 | **58** | 1030 | **53** |
| DCCA | 56 | 627 | 79 | 58 | 516 | 90 | 21 | 2095 | 913 |
| GCN | 60 | 546 | 45 | 62 | 485 | 71 | 54 | 1130 | 108 |
| Text+Network (transductive) | | | | | | | | | |
| MLP-TXT+NET 1% | **8** | 1521 | 1295 | 14 | 1436 | 1411 | 8 | 3865 | 2041 |
| DCCA 1% | 7 | 1425 | 979 | 38 | 869 | 348 | 14 | 3014 | 1367 |
| GCN 1% | 6 | **1103** | **609** | **41** | **788** | **311** | **21** | **2071** | **853** |

Table 1: Geolocation results over the three Twitter datasets for the proposed models: joint text+network MLP-TXT+NET, DCCA, and GCN and network-based GCN-LP. The models are compared with text-only and network-only methods. The performance of the three joint models is also reported for minimal supervision scenario where only 1% of the total samples are labelled. "—" signifies that no results were reported for the given metric or dataset. Note that Do et al. (2017) use timezone, and Miura et al. (2017) use the description and location fields in addition to text and network.

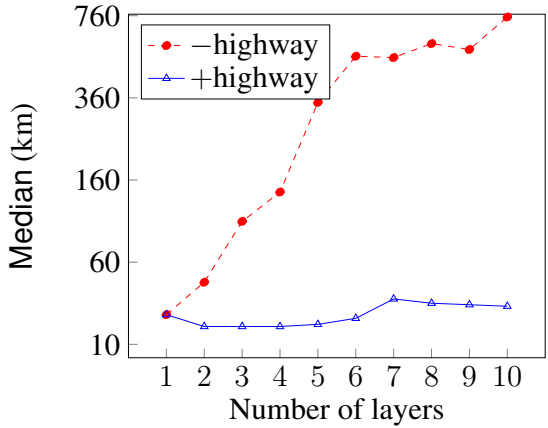

Figure 5: The effect of adding more GCN layers (neighbourhood expansion) to GCN in terms of median error over the development set of GEOTEXT with and without the highway gates, and averaged over 5 runs.

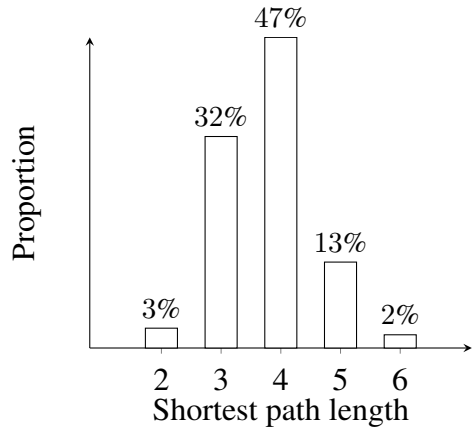

Figure 6: The distribution of shortest path lengths between all the nodes of the largest connected component of GEOTEXT's graph that constitute more than 1% of total.

TEXT. However, it's difficult to make a fair comparison as they use timezone data in their feature set. MLP-TXT+NET outperforms GCN over TWITTER-US and TWITTER-WORLD, which are very large, and have large amounts of labelled data. In a scenario with little supervision (1% of the total samples are labelled) DCCA and GCN clearly outperform MLP-TXT+NET, as they have fewer pa-rameters. Except for Acc@161 over GEOTEXT where the number of labelled samples in the minimal supervision scenario is very low, GCN outperforms DCCA by a large margin, indicating that for a medium dataset where only 1% of samples are labelled (as happens in random samples of Twitter) GCN is superior to MLP-TXT+NET and DCCA, consistent with Section 4.2. Both MLP-TXT+NET and GCN achieve state of the art results compared to network-only, text-only, and hybrid models. The network-based `GCN-LP` model, which does label propagation using Graph Convolutional Networks, outperforms Rahimi et al. (2015a), which is based on location propagation using Modified Adsorption (Talukdar and Crammer, 2009), possibly because the label propagation in `GCN` is parametrised.

### 4.5 Error Analysis

Although the performance of `MLP-TXT+NET` is better than `GCN` and `DCCA` when a large amount of labelled data is available (Table 1), under a scenario where little labelled data is available (1% of data), `DCCA` and `GCN` outperform `MLP-TXT+NET`, mainly because the number of parameters in `MLP-TXT+NET` grows with the number of samples, and is much larger than `GCN` and `DCCA`. `GCN` outperforms `DCCA` and `MLP-TXT+NET` using 1% of data, however, the distribution of errors in the development set of TWITTER-US indicates higher error for smaller states such as Rhode Island (RI), Iowa (IA), North Dakota (ND), and Idaho (ID), which is simply because the number of labelled samples in those states is insufficient.

Although we evaluate geolocation models with `Median`, `Mean`, and `Acc@161`, it doesn't mean that the distribution of errors is uniform over all locations. Big cities often attract more local online discussions, making the geolocation of users in those areas simpler. For example users in LA are more likely to talk about LA-related issues such as their sport teams, Hollywood or local events than users in the state of Rhode Island (RI), which lacks large sport teams or major events. It is also possible that people in less densely populated areas are further apart from each other, and therefore, as a result of discretisation fall in different clusters. The non-uniformity in local discussions results in lower geolocation performance in less densely populated areas like Midwest U.S., and higher performance in densely populated areas such as NYC and LA as shown in Figure 7. The geographical distribution of error for `GCN`, `DCCA` and `MLP-TXT+NET` under the minimal supervision scenario is shown in the supplementary material.

To get a better picture of misclassification between states, we built a confusion matrix based on known state and predicted state for development users of TWITTER-US using `GCN` using only 1% of labelled data. There is a tendency for users to be wrongly predicted to be in CA, NY, TX, and surprisingly OH. Particularly users from states such as TX, AZ, CO, and NV, which are located close to CA, are wrongly predicted to be in CA, and users from NJ, PA, and MA are misclassified as being in NY. The same goes for OH and TX where users from neighbouring smaller states are misclassified to be there. Users from CA and NY are also misclassified between the two states, which might be the result of business and entertainment connections that exist between NYC and LA/SF. Interestingly, there are a number of misclassifications to FL for users from CA, NY, and TX, which might be the effect of users vacationing or retiring to FL. The full confusion matrix between the U.S. states is provided in the supplementary material.

### 4.6 Local Terms

In Table 2, local terms of a few regions detected by `GCN` under minimal supervision are shown. The terms that were present in the labelled data are excluded to show how graph convolutions over the social graph have extended the vocabulary. For example, in case of Seattle, *#goseahawks* is an important term not present in the 1% labelled data but present in the unlabelled data. The convolution over the social graph is able to utilise such terms that don't exist in the labelled data.

## 5 Related Work

Previous work on user geolocation can be broadly divided into text-based, network-based and multi-view approaches.

Text-based geolocation uses the geographical bias in language use to infer the location of users. There are three main text-based approaches to geolocation: (1) gazetteer-based models which map geographical references in text to location, but ignore non-geographical references and vernacular uses of language (Rauch et al., 2003; Amitay et al., 2004; Lieberman et al., 2010); (2) geographical topic models that learn region-specific topics, but don't scale to the magnitude of social media (Eisenstein et al., 2010; Hong et al., 2012; Ahmed et al., 2013); and (3) supervised models which are often framed as text classification (Serdyukov et al., 2009; Wing and Baldridge, 2011; Roller et al., 2012; Han et al., 2014) or text regression (Iso et al., 2017; Rahimi et al., 2017a). Supervised models scale well and can achieve good performance with sufficient supervision, which is not available in a real world scenario.

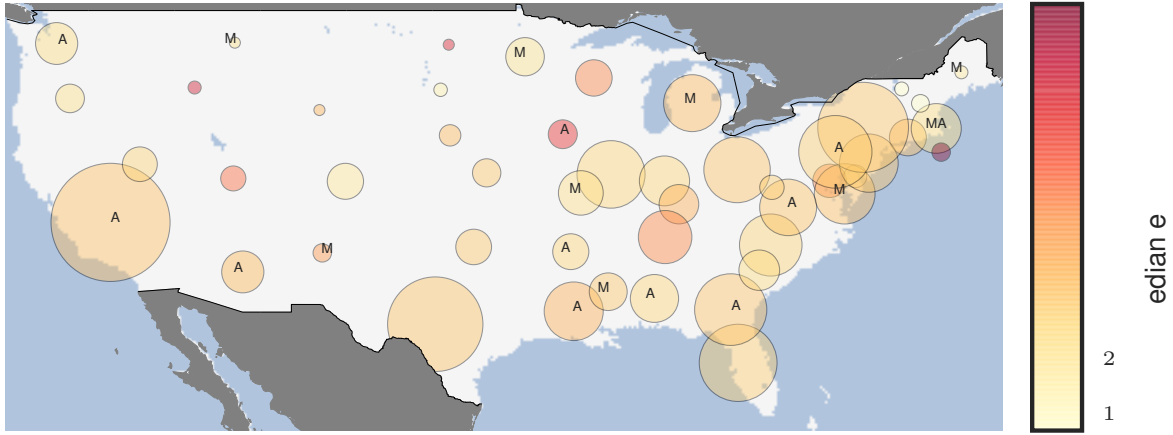

Figure 7: The geographical distribution of Median error of GCN using 1% of labelled data in each state over the development set of TWITTER-US. The colour indicates error and the size indicates the number of development users within the state.

| Seattle, WA | Austin, TX | Jacksonville, FL | Columbus, OH | Charlotte, NC | Phoenix, AZ | New Orleans, LA | Baltimore, MD |
|---|---|---|---|---|---|---|---|
| #goseahawks | stubb | unf | laffayette | #asheville | clutterbuck | mcneese | bhop |
| smock | gsd | ribault | #weareohio | #depinga | waffels | keela | #dsu |
| traffuck | #meatsweats | wahoowa | #arcgis | batesburg | bahumbug | pentecostals | chestertown |
| ferran | lanterna | wjct | #slammin | stewey | iedereen | lutcher | aduh |
| promissory | pupper | fscj | #ouhc | #bojangles | rockharbor | grogan | umbc |
| chowdown | effaced | floridian | #cow | #occupyraleigh | redtail | suela | lmt |
| ckrib | #austin | #jacksonville | mommyhood | gville | gewoon | cajuns | assistly |
| #uwhuskies | lmfbo | #mer | beering | sweezy | jms | bmu | slurpies |

Table 2: Top terms for selected regions detected by GCN using only 1% of TWITTER-US for supervision. We present the terms that were present only in unlabelled data. The terms include city names, hashtags, food names and internet abbreviations.

Network-based methods leverage the *location homophily* assumption: nearby users are more likely to befriend and interact with each other. There are four main network-based geolocation approaches: distance-based, supervised classification, graph-based label propagation, and node embedding methods. Distance-based methods model the probability of friendship given the distance (Backstrom et al., 2010; McGee et al., 2013; Gu et al., 2012; Kong et al., 2014), supervised models use neighbourhood features to classify a user into a location (Rout et al., 2013; Malmi et al., 2015), and graph-based label-propagation models propagate the location information through the user–user graph to estimate unknown labels (Davis Jr et al., 2011; Jurgens, 2013; Compton et al., 2014). Node embedding methods build heterogeneous graphs between user–user, user–location and location–location, and learn an embedding space to minimise the distance of connected nodes, and maximise the distance of disconnected nodes. The embeddings are then used in supervised models for geolocation (Wang et al., 2017). Network-based models fail to geolocate disconnected users: Jurgens et al. (2015) couldn't geolocation 37% of users as a result of disconnectedness.

Previous work on hybrid text and network methods can be broadly categorised into three main approaches: (1) incorporating text-based information such as toponyms or locations predicted from a text-based model as auxiliary nodes into the user–user graph, which is then used in network-based models (Li et al., 2012a,b; Rahimi et al., 2015b,a); (2) ensembling separately trained text- and network-based models (Gu et al., 2012; Ren et al., 2012; Jayasinghe et al., 2016; Ribeiro and Pappa, 2017); and (3) jointly learning geolocation from several information sources such as text and network information (Miura et al., 2017; Do et al., 2017), which can capture the complementary information in text and network views, and also model the interactions between the two. None of the previous multiview approaches — with the exception of Li et al. (2012a) and Li et al. (2012b) that only use toponyms — effectively uses unlabelled data in the text view, and use only the unlabelled information of the network view via the user–user graph.

There are three main shortcomings in the previous work on user geolocation that we address in this paper: (1) with the exception of few recent works (Miura et al., 2017; Do et al., 2017), previous models don't jointly exploit both text and network information, and therefore the interaction between text and network views is not modelled; (2) the unlabelled data in both text and network views is not effectively exploited, which is crucial given the small amounts of available supervision; and (3) previous models are rarely evaluated under a minimal supervision scenario, a scenario which reflects real world conditions.

## 6 Conclusion

We proposed `GCN`, `DCCA` and `MLP-TXT+NET`, three multiview, transductive, semi-supervised geolocation models, which use text and network information to infer user location in a joint setting. We showed that joint modelling of text and network information outperforms network-only, text-only, and hybrid geolocation models as a result of modelling the interaction between text and network information. We also showed that `GCN` and `DCCA` are able to perform well under a minimal supervision scenario similar to real world applications by effectively using unlabelled data. We ignored the context in which users interact with each other, and assumed all the connections to hold location homophily. In future work, we are interested in modelling the extent to which a social interaction is caused by geographical proximity (e.g. using user–user gates).